\documentclass{article}
\usepackage{spconf,amsmath,graphicx}
\usepackage{multirow}
\usepackage{tabularray}
\usepackage{subcaption}
\usepackage[shortlabels]{enumitem}
\usepackage[hidelinks]{hyperref}
\usepackage{indentfirst}
\usepackage{booktabs}
\usepackage{arydshln}
\usepackage{dcolumn}
\usepackage[table]{xcolor}
\usepackage{amssymb}
\usepackage{cite}

\title{Survey on Single-Image Reflection Removal using Deep Learning Techniques}
\name{Kangning Yang, Huiming Sun, Jie Cai, Lan Fu, Jiaming Ding, Jinlong Li, Chiu Man Ho, Zibo Meng}
\address{OPPO AI Center}

\begin{document}
%
\maketitle
\begin{abstract}
The phenomenon of reflection is quite common in digital images, posing significant challenges for various applications such as computer vision, photography, and image processing. Traditional methods for reflection removal often struggle to achieve clean results while maintaining high fidelity and robustness, particularly in real-world scenarios. Over the past few decades, numerous deep learning-based approaches for reflection removal have emerged, yielding impressive results. In this survey, we conduct a comprehensive review of the current literature by focusing on key venues such as ICCV, ECCV, CVPR, NeurIPS, etc., as these conferences and journals have been central to advances in the field. Our review follows a structured paper selection process, and we critically assess both single-stage and two-stage deep learning methods for reflection removal. The contribution of this survey is three-fold: first, we provide a comprehensive summary of the most recent work on single-image reflection removal; second, we outline task hypotheses, current deep learning techniques, publicly available datasets, and relevant evaluation metrics; and third, we identify key challenges and opportunities in deep learning-based reflection removal, highlighting the potential of this rapidly evolving research area.
\end{abstract}

\section{Introduction}
\label{sec:intro}
Single-image reflection removal (SIRR) is a critical task in image processing, focusing on recovering the true scene behind reflections from reflective surfaces (e.g., transparent glasses). 

Over the years, various techniques have been proposed to solve the SIRR problem. Traditional methods typically relied on a non-learning paradigm. Since the goal of SIRR is to recover the transmission image (i.e., the true scene) from a blended image containing both the scene and reflections, this is an ill-posed problem. In the absence of additional information about the scene, there are an infinite number of possible decompositions of the image~\cite{levin2004separating}. Therefore, these traditional methods often exploit prior knowledge (or priors) to constrain the solution space and guide the recovery process. One widely used prior is sparsity priors~\cite{levin2002learning, levin2007user}. These approaches impose gradient sparsity constraints to find the minimum edges and corners of the layer decomposition. The fundamental idea is to require the image gradient histogram to have a long-tail distribution~\cite{li2014single}. Another common assumption is the use of smoothness priors~\cite{li2014single, yang2019fast, wan2016depth}. This prior is based on the observation that reflection layers are more likely to be blurred compared to the background scene, primarily due to differences in the distance from the camera. Additionally, Shih et al.~\cite{shih2015reflection} introduced the concept of examining ghosting effects caused by double-pane windows. These ghosting effects lead to multiple reflections in the captured image, which can be challenging to separate. To address this, they proposed modeling these effects using Gaussian Mixture Models (GMM) as a patch-based prior.

However, these priors often struggle to generalize well across different types of reflections and scenes, particularly when dealing with complex real-world environments. Most of these priors are based on the assumption that the captured image $I$ is a linear combination of the transmitted scene $T$ and a reflection $R$, i.e., $I = T + R$. While this assumption is simple, it often deviates significantly from reality, as real-world reflections are more complex and influenced by factors such as lighting conditions, surface characteristics, and so on~\cite{song2023robust}. Moreover, both $T$ and $R$ may contain content from real-world scenes, leading to overlapping appearance distributions that make the separation of the two components more challenging~\cite{wei2019single}. To address these issues, researchers have recently shifted focus toward learning-based methods, particularly data-driven deep learning approaches, driven by the success of deep neural networks in tackling versatile computer vision problems. Instead of relying on handcrafted priors, deep learning models can be trained on large, labeled datasets, enabling them to effectively handle a broad spectrum of scenarios. 

This article presents a comprehensive survey of SIRR research utilizing deep learning methods. We analyze current research trends and identify future opportunities. Compared to existing surveys on similar topics, this article offers a more thorough review of the literature, focusing on key journals and conferences, with the aim to of presenting concise, critical, and recent research advances. For example, although Wan et al.\cite{wan2017benchmarking, wan2022benchmarking} provided a brief survey, their focus was primarily on introducing their new datasets (SIR$^2$ and SIR$^{2+}$) and establishing benchmarks for different algorithms. Amanlou et al.\cite{amanlou2022single} conducted a survey on SIRR using deep learning, but their coverage is limited to work published between 2015 and 2021. In contrast, our review covers the most recent advancements and organizes the research within a more detailed framework.

The article is organized as follows: Section 2 describes the methodology used to conduct the bibliographic search. Section 3 introduces the mathematical hypotheses for modeling SIRR. Section 4 surveys the SIRR research from three perspectives: single-stage, two-stage, and multi-stage approaches. Section 5 discusses currently available public datasets and commonly used evaluation metrics. Finally, Section 6 explores future research opportunities in SIRR field, followed by brief concluding remarks in Section 7.
\section{Methodology}
\label{sec:methodology}
To conduct a comprehensive and focused survey of the relevant literature, we strategically concentrated our bibliographic search on key conferences and journals that are widely recognized for publishing state-of-the-art high-impact research in computer vision and artificial intelligence. Instead of conducting a broad search across all available digital libraries, we specifically targeted prominent venues considered the most influential in these domains. These include CVPR, ICCV, ECCV, WACV, NeurIPS, TPAMI, TIP, and Appl. Intell.. Papers presented at these conferences and published in these journals are widely regarded for their high impact, having undergone rigorous peer review and earned broad acknowledgment within the academic community. The high citation counts of works from these venues further demonstrate their significant influence and acceptance, ensuring that the studies included in our survey are relevant and credible.

We used the following search query to capture relevant research on SIRR:  [(``single image reflection removal" OR ``single image reflection erase" OR ``single image reflection separation" OR ``single image reflection elimination" OR ``single-image reflection removal" OR ``single-image reflection erase" OR ``single-image reflection separation" OR ``single-image reflection elimination") AND (``deep learning" or ``neural network" OR ``artificial intelligence")]. We excluded papers that were not full-text research articles, such as tutorials, abstracts, workshops, posters, etc. We also applied a time filter in order to consider only publications between 2017 and 2025 (as Fan et al.~\cite{fan2017generic} introduced the first neural network model for solving SIRR in 2017). We then analyzed each of the papers to ensure they were appropriate for this survey. Finally, 28 papers remained which constitute the basis of our literature review.
\section{Mathematical Hypothesis}
\label{sec:hypothesis}
As previously mentioned, SIRR is inherently an ill-posed problem. To this end, researchers have proposed various hypotheses.

\subsection{Linear Hypothesis}
The linear hypothesis posits that a captured image $I$ is perceived as the superimposition of a transmission layer and a reflection layer, a concept inspired by the human visual system~\cite{levin2002learning}.

Early studies, mainly non-learning approaches~\cite{li2014single, levin2007user} and early deep learning works~\cite{zhang2018single, fan2017generic, hu2021trash}, adopted this hypothesis, assuming that an image containing reflections can be mathematically modeled as the sum of the transmission and reflection layers, i.e., $I = T + R$. However, this assumption is heuristic; the reflection and transmission layers are likely to degrade due to diffusion during the superposition process~\cite{hu2023single, wan2020reflection}, and it may not hold true in cases involving intense and bright reflections~\cite{li2023two}.

Furthermore, researchers~\cite{wan2018crrn, yang2018seeing, li2020single} introduced blending scalars to build a more nuanced model, $I = \alpha T + \beta R$, where $\alpha$ and $\beta$ represent scaling factors for the transmission and reflection layers, respectively. For instance, Li et al.~\cite{li2020single} assumes $I = \alpha T + R$, while Yang et al.~\cite{yang2018seeing} assumes that $I = \alpha T + (1-\alpha)R$, and Wan et al.~\cite{wan2018crrn} treats $\alpha$ and $\beta$ as mixing coefficients balancing the transmission and reflection layers. 

Despite these improvements, blending two images using constant values does not accurately simulate the complex real-world reflection process. The formation of the reflected image depends on factors such as the relative position of the camera to the image plane and the lighting conditions~\cite{wen2019single}.

\subsection{Non-linear Hypothesis}
To address the complex process of image reflection, some studies have leveraged the full potential of deep learning to efficiently incorporate prior information mined from labeled training data into network structures. These studies also introduced non-linearity to develop more sophisticated models that better approximate the physical mechanisms involved in image formation~\cite{dong2021location, li2020single, wei2019single}. One such approach utilizes alpha matting~\cite{dong2021location} to model the blending process. In this formulation, an alpha blending mask $W$ is introduced to represent the relative contribution of the transmission layer at each pixel. The synthesis process is then represented as: $I = W \circ T + R$, where $\circ$ denotes the element-wise multiplication. Similarly, Wen et al.~\cite{wen2019single} approximates the reflection mechanisms as: $I = W \circ T + (1 - W) \circ R$. This approach enables a more flexible and accurate separation of the scene and reflection layers, particularly in cases involving complex light interactions or semi-transparent surfaces. 

Additionally, Zheng et al.~\cite{zheng2021single} proposed the model $I = \Omega T + \Phi R$, where $\Omega$ and $\Phi$ represent refractive and reflective amplitude coefficient maps, respectively. Likewise, Wan et al.~\cite{wan2020reflection} considered degradation in both $T$ and $R$, expressing the formation of a mixed image as $I = g(T) + f(R)$, where $g(\cdot)$ and $f(\cdot)$ represent the various degradation processes learned by the network structure. More recently, Hu and Guo~\cite{hu2023single} deliver a more general formulation as: $I = T + R + \Phi(T, R)$, where $\Phi(T, R)$ represents the residue in the reconstruction process, which may arise due to factors such as attenuation, overexposure, etc.

\begin{table*}
\centering
\renewcommand{\dashlinedash}{0.5pt} 
\renewcommand{\dashlinegap}{2pt}    
\begin{tabular}{lllll}
\hline\hline
 & \shortstack[l]{\space \\ \space \\ \textbf{Methods} \\ \space } & 
   \shortstack[l]{\space \\ \space \\ \textbf{Venue} \\ \space } & 
   \shortstack[l]{\space \\ \space \\ \textbf{Scheme} \\ \space } & 
   \shortstack[l]{\space \\ \space \\ \textbf{Cross-stage fusion} \\ \space} \\
\bottomrule   \\
Single-stage
& Zhang et al.~\cite{zhang2018single} & CVPR 2018 & $I \rightarrow [T, R]$ & - \\ 
& ERRNet~\cite{wei2019single} & CVPR 2019 & $I \rightarrow T$ & - \\ 
& RobustSIRR~\cite{song2023robust} & CVPR 2023 & $I_{multiscale} \rightarrow T$ & - \\
& YTMT~\cite{hu2021trash} & NeurIPS 2021 & $I \rightarrow [T, R]$ & - \\   \bottomrule   \\
Two-stage
& CoRRN~\cite{wan2019corrn} & TPAMI 2019 & \shortstack[l]{\space \\$I \rightarrow E_{T}$ \\ $[I, E_{T}] \rightarrow T$} & Convolutional Fusion \\ \cline{2-5}
& DMGN~\cite{feng2021deep} & TIP 2021 & \shortstack[l]{\space \\ $I \rightarrow [T_{1}, R]$ \\ $[I, T_{1}, R] \rightarrow T$} & Convolutional Fusion \\ \cline{2-5}
& RAGNet~\cite{li2023two} & Appl. Intell. 2023 & \shortstack[l]{\space \\ $I \rightarrow R$ \\ $ [I, R] \rightarrow T$} & Convolutional Fusion \\ \cline{2-5}
& CEILNet~\cite{fan2017generic} & ICCV 2017 & \shortstack[l]{\space \\ $[I, E_{I}] \rightarrow E_{T}$ \\ $[I, E_{T}] \rightarrow T$ } & Concat \\ \cline{2-5}
& DSRNet~\cite{hu2023single} & ICCV 2023 & \shortstack[l]{\space \\ $I \rightarrow (T_{1}, R_{1})$ \\ $(R_{1}, T_{1}) \rightarrow (R, T, residue)$} & N/A \\ \cline{2-5}
& SP-net BT-net~\cite{kim2020single} & CVPR 2020 & \shortstack[l]{\space \\ $I \rightarrow [T_{1}, R_{1}]$ \\ $R_{1} \rightarrow R$} & N/A \\ \cline{2-5}
& Wan et al.~\cite{wan2020reflection} & CVPR 2020  & \shortstack[l]{\space \\ $[I, E_{I}] \rightarrow R_{1}$ \\ $R_{1} \rightarrow R$} & N/A \\ \cline{2-5}
& Zheng et al.~\cite{zheng2021single} & CVPR 2021 & \shortstack[l]{\space \\ $I \rightarrow e$ \\ $[I, e] \rightarrow T$} & Concat \\ \cline{2-5}
& Zhu et al.~\cite{zhu2024revisiting} & CVPR 2024 & \shortstack[l]{\space \\ $I \rightarrow E_{R}$ \\ $ [I, E_{R}] \rightarrow T$} & Concat \\ \cline{2-5}
& Language-Guided~\cite{zhong2024language} & CVPR 2024 & \shortstack[l]{\space \\ $[I, Texts] \rightarrow R\ or\ T$ \\ $[I, R\ or\ T] \rightarrow T\ or\ R$} & Feature-Level Concat
\\   \bottomrule   \\
Multi-stage
& BDN~\cite{yang2018seeing} & ECCV 2018 & \shortstack[l]{\space \\ $I \rightarrow T_{1}$ \\ $[I, T_{1}] \rightarrow R$ \\ $[I, R] \rightarrow T$ } & Concat \\ \cline{2-5}
& IBCLN~\cite{li2020single} & CVPR 2020 & \shortstack[l]{\space \\ $[I, R_{0}, T_{0}] \rightarrow [R_{1}, T_{1}]$ \\ $[I, R_{1}, T_{1}] \rightarrow [R_{2}, T_{2}]$ \\ \ldots} & \shortstack[l]{Concat \\ Recurrent} \\ \cline{2-5}
& Chang et al.~\cite{chang2021single} & WACV 2021 & \shortstack[l]{\space \\ $I \rightarrow E_{T}$ \\ $[I, E_{T}] \rightarrow T_{1} \rightarrow R_{1} \rightarrow T_{2}$ \\ $[I, E_{T}, T_{2}] \rightarrow R \rightarrow T$} & \shortstack[l]{Concat \\ Recurrent} \\ \cline{2-5}
& LANet~\cite{dong2021location} & ICCV 2021 & \shortstack[l]{\space \\ $[I, T_{0}] \rightarrow R_{1} \rightarrow T_{1}$ \\ $[I, T_{1}] \rightarrow R_{2} \rightarrow T_{2}$ \\ \ldots} & \shortstack[l]{Concat \\ Recurrent} \\ \cline{2-5}
& V-DESIRR~\cite{prasad2021v} & ICCV 2021 & \shortstack[l]{\space \\ $I_{1} \rightarrow T_{1}$ \\ $[I_{1}, T_{1}, I_{2}] \rightarrow T_{2}$ \\ \ldots \\ $[I_{n-1}, T_{n-1}, I_{n}] \rightarrow T$ } & \shortstack[l]{Convolutional Fusion \\ Recurrent} \\
\hline\hline
\end{tabular}
\caption{\textbf{I}, \textbf{R}, \textbf{T}, and \textbf{E} represent the \textbf{I}nput, \textbf{R}eflection, \textbf{T}ransmission, and \textbf{E}dge map, respectively. The subscripts of \textbf{T} and \textbf{R} represent intermediate process outputs. The Absorption Effect $e$ is introduced in~\cite{zheng2021single} to describe light attenuation as it passes through the glass. The output $residue$ term, proposed in~\cite{hu2023single}, is used to correct errors in the additive reconstruction of the reflection and transmission layers. Language descriptions in~\cite{zhong2024language} provide contextual information about the image layers, assisting in addressing the ill-posed nature of the reflection separation problem.}
\label{tab:1}
\end{table*}

\section{Reflection Removal Approaches}
We first present one-stage SIRR approaches in Section~\ref{sec:single-stage}, followed by two-stage and multi-stage approaches in Sections~\ref{sec:two-stage} and~\ref{sec:multi-stage}. The learning objective is introduced in Section~\ref{sec:learning-objective}, with a comparative analysis of models in Table~\ref{tab:1}.

\subsection{Single-Stage Approaches}
\label{sec:single-stage}

In the field of reflection removal, most academic approaches adopt a multi-stage architecture. However, some studies have also proposed one-stage architectures. Given an image \( I \in [0, 1]^{m \times n \times 3} \) with reflections, these approaches typically decompose \( I \) into a transmission layer \( f_T(I; \theta) \) and/or a reflection layer \( f_R(I; \theta) \), using a single network such that $I \approx f_T(I; \theta) + f_R(I; \theta)$, where \( \theta \) represents the network parameters.
ERRNet~\cite{wei2019single} and RobustSIRR~\cite{song2023robust} take \( I \) as input and output only \( T \), while Zhang et al.~\cite{zhang2018single} and YTMT~\cite{hu2021trash} take \( I \) as input and output both \( R \) and \( T \). Additionally, RobustSIRR~\cite{song2023robust} utilizes multi-resolution inputs alongside to enhance feature extraction and improve reflection removal performance.

Zhang et al.~\cite{zhang2018single} propose utilizing a deep neural network with perceptual losses to address the problem of SIRR. ERRNet~\cite{wei2019single} enhances a fundamental image reconstruction neural network by simplifying residual blocks through the elimination of batch normalization, expanding capacity by widening the network to 256 feature maps, and enriching the input image \( I \)  with hypercolumn features extracted from a pretrained VGG-19 network to incorporate semantic information for better performance. RobustSIRR~\cite{song2023robust} presents a robust transformer-based model for SIRR, integrating cross-scale attention modules, multi-scale fusion modules, and an adversarial image discriminator to improve performance. YTMT~\cite{hu2021trash} introduces a simple yet effective interactive strategy called "Your Trash is My Treasure". This approach constructs dual-stream decomposition networks by facilitating block-wise communication between the streams and transferring deactivated ReLU information from one stream to the other, leveraging the additive property of the components.

\subsection{Two-Stage Approaches}
\label{sec:two-stage}

Due to the inherent ambiguity and complexity of SIRR, solving this problem is very challenging. To address this, researchers typically use deep learning models in a cascade or sequence manner, which helps manage the uncertainty in estimating transmission layer while also simplifying the training of SIRR systems.
Some academic approaches adopt a two-stage architecture, where an intermediate output, such as a reflection layer~\cite{feng2021deep,li2023two,hu2023single,kim2020single,wan2020reflection,zhong2024language}, a coarse transmission layer~\cite{feng2021deep,hu2023single,kim2020single,zhong2024language}, an edge map~\cite{wan2019corrn,fan2017generic,zhu2024revisiting}, and so on~\cite{zheng2021single}, is first estimated, followed by the reconstruction of the final transmission layer and/or reflection layer. Besides, the techniques for fusing features across different stages vary among studies. Some methods use convolutional fusion techniques~\cite{wan2019corrn, feng2021deep, li2023two}, while others apply image-level~\cite{fan2017generic, zheng2021single, zhu2024revisiting} or feature-level concatenation~\cite{zhong2024language}. Additionally, certain approaches~\cite{hu2023single, kim2020single, wan2020reflection} do not employ any fusion strategies.

CoRRN\cite{wan2019corrn} proposes a network that uses feature sharing to tackle the problem within a cooperative framework, combining image context and multi-scale gradient information. 
DMGN~\cite{feng2021deep} presents a unified framework for background restoration, employing the Residual Deep-Masking Cell to progressively refine and control information flow. 
The RAG~\cite{li2023two} module is designed to improve the use of the estimated reflection for more accurate transmission layer prediction. 
CEILNet~\cite{fan2017generic} introduces a cascaded pipeline for edge prediction followed by image reconstruction.
DSRNet~\cite{hu2023single} architecture features two cascaded stages and a learnable residue module (LRM). Stage 1 gathers hierarchical semantic information, while Stage 2 refines the decomposition using the LRM to separate components that break the linear assumption.
The structure in~\cite{kim2020single} includes SP-net, which decomposes an input image into the predicted transmission layer and background reflection layer. The BT-net then eliminates glass and lens effects from the predicted reflection, enhancing image clarity and enabling more accurate error matching.
In this paper~\cite{wan2020reflection}, instead of eliminating reflection components from the mixed image, the goal is to recover the reflection scenes from the mixture.
This paper~\cite{zheng2021single} addresses SIRR by incorporating the absorption effect $(e)$, which is approximated using the average refractive amplitude coefficient map. It proposes a two-step solution: the first step estimates the absorption effect from the reflection-contaminated image, and the second step recovers the transmission image using both the reflection-contaminated image and the estimated absorption effect.
The framework~\cite{zhu2024revisiting} consists of RDNet and RRNet, where RDNet utilizes a pretrained backbone with residual blocks and interpolation to estimate the reflection mask, and RRNet uses this estimate to assist in the reflection removal process.
This paper~\cite{zhong2024language} addresses language-guided reflection separation by using language descriptions to provide layer content. It proposes a unified framework that uses cross-attention and contrastive learning to align language descriptions with image layers, while a gated network and randomized training strategy help resolve layer ambiguity.

\subsection{Multi-Stage Approaches}
\label{sec:multi-stage}

Some studies extend beyond a two-stage architecture by using a multi-stage cascaded structure. Similar to the two-stage design, the multi-stage approach generates intermediate outputs in a recurrent fashion, eventually reconstructing the final transmission and/or reflection layer.
Some methods use convolutional fusion techniques~\cite{prasad2021v}, while others utilize concatenation~\cite{yang2018seeing, li2020single, chang2021single, dong2021location}.

DBN~\cite{yang2018seeing} introduces a cascaded deep neural network that simultaneously estimates both background and reflection components. The network follows a bidirectional approach: first using the estimated background to predict the reflection, and then refining the background prediction using the estimated reflection. This dual-estimation strategy improves reflection removal performance.
IBCLN~\cite{li2020single} is designed for reflection removal by progressively refining the estimates of the transmission and reflection layers, with each iteration improving the prediction of the other. By utilizing LSTM to transfer information between steps and incorporating residual reconstruction loss, IBCLN tackles the vanishing gradient issue and improves training across multiple cascade steps.
The model in~\cite{chang2021single} takes a reflection-contaminated image and separates it into the reflection and transmission layers. To ensure high-quality transmission, three auxiliary techniques are employed: edge guidance, a reflection classifier, and recurrent decomposition.
This paper~\cite{dong2021location} presents a LANet for SIRR. It employs a reflection detection module that generates a probabilistic confidence map using multi-scale Laplacian features. The network, designed as a recurrent model, progressively refines reflection removal, with Laplacian kernel parameters highlighting strong reflection boundaries to improve detection and enhance the quality of the results.
V-DESIRR~\cite{prasad2021v} introduces a lightweight model for reflection removal using an innovative scale-space architecture, which processes the corrupted image in two stages: a Low Scale Sub-network (LSSNet) for the lowest scale and a Progressive Inference (PI) stage for higher scales. To minimize computational complexity, the PI stage sub-networks are significantly shallower than LSSNet, and weight sharing across scales enables the model to generalize to high resolutions without the need for retraining.

\subsection{Learning Objective}
\label{sec:learning-objective}

To train SIRR models, several commonly used loss functions are combined to ensure high-quality reflection removal. These include Reconstruction Loss, Perceptual Loss\cite{johnson2016perceptual}, and Adversarial Loss\cite{goodfellow2014generative}. Each of these loss functions contributes to different aspects of the model’s learning process:

\textbf{Reconstruction loss} is typically defined using the $L1$ or $L2$ loss, which directly measures the pixel-wise difference between the predicted reflection-free image and the ground truth image. This loss ensures that the output image is as close as possible to the desired reflection-free image in a pixel-wise sense. However, relying solely on this loss can lead to overly smooth results, as it does not consider high-level perceptual differences. The $L1$ loss formulation is as follows:

\begin{equation}
    \mathcal{L}_{\text{rec}} = \| \hat{T} - T \|_1
\end{equation}

where \( \hat{T} \) is the predicted reflection-free image and \( T \) is the ground truth image.

To mitigate the oversmoothing effect of reconstruction loss and preserve important structural details, \textbf{Gradient Consistency Loss} is utilized in~\cite{wei2019single}. This loss ensures that the predicted transmission layer \( \hat{T} \) retains the edge structures of the ground truth \( T \) by minimizing the difference between their gradients along both the \( x \)- and \( y \)-directions:

\begin{equation}
    \mathcal{L}_{\text{grad}} = \|\nabla_x \hat{T} - \nabla_x T\|_1 + \|\nabla_y \hat{T} - \nabla_y T\|_1
\end{equation}

where \( \nabla_x \) and \( \nabla_y \) are the gradient operators along the horizontal and vertical directions, respectively.

By combining these two losses, SIRR achieves a balance between accurate pixel-wise reconstruction and the preservation of structural details.

\textbf{Perceptual loss} utilizes a pre-trained deep neural network (e.g., VGG) to extract high-level feature representations of both the predicted and ground truth images. Instead of measuring pixel-wise differences, this loss compares the differences in feature space, making the generated images more visually realistic and closer to human perception. The perceptual loss can be expressed as:

\begin{equation}
    \mathcal{L}_{\text{per}} = \sum_{i} \| \phi_i(\hat{T}) - \phi_i(T) \|_1
\end{equation}

where \( \phi_i(\cdot) \) represents the feature map extracted from the \( i \)-th layer of the pre-trained network.

\textbf{Adversarial loss} is inspired by Generative Adversarial Networks (GANs) and is used to improve the realism of the generated images. A discriminator \( D \) is introduced to distinguish between real reflection-free images and the generated images. The adversarial loss is formulated as:

\begin{equation}
    \mathcal{L}_{\text{adv}} = \mathbb{E}[\log D(T)] + \mathbb{E}[\log (1 - D(\hat{T}))]
\end{equation}

where \( D(\cdot) \) represents the discriminator network. The generator aims to minimize this loss, making the generated images indistinguishable from real ones.

In addition to reconstruction loss, perceptual loss, and adversarial loss, other loss functions are also utilized to enhance reflection removal performance. One such example is the \textbf{exclusion loss}~\cite{li2023two,zhang2018single}, which encourages the separation of the transmission (\(T\)) and reflection (\(R\)) layers by minimizing their structural correlation. This loss enforces gradient decorrelation at multiple scales. It is formulated as:

\begin{equation}
    \mathcal{L}_{\text{excl}} = \frac{1}{N+1} \sum_{n=0}^{N} \sqrt{\|\Psi(T\downarrow^n, R\downarrow^n)\|_F}
\end{equation}

where \(T\downarrow^n\) and \(R\downarrow^n\) are the downsampled versions at different scales, $\|\cdot\|_F$ is the Frobenius norm, and $\Psi(T, R)$ measures the correlation between their gradients.




\textbf{Total Variation Loss} is a regularization technique commonly used in image processing tasks to promote smoothness and reduce noise or artifacts. It encourages spatial continuity by minimizing the differences between neighboring pixels, preventing excessive sharp variations. TVLoss is particularly useful in reflection removal~\cite{zhu2024revisiting}, denoising, and super-resolution tasks, where it helps generate cleaner and more visually appealing results by reducing undesired texture artifacts while preserving important image structures.

\textbf{Contextual Loss} is used in~\cite{prasad2021v} to preserve fine-grained details in image generation tasks by focusing on feature similarity rather than direct pixel-wise differences. The formula for Contextual Loss is typically expressed as:

\[
\mathcal{L}_{\text{CX}}(F, F^*) = - \log \left( \max_j \ \text{CX}_{ij} \right)
\]

where \( F \) and \( F^* \) represent feature activations from a pre-trained network for the generated and target images, respectively. The contextual similarity \( \text{CX}_{ij} \) measures the correlation between feature vectors, ensuring that the generated image retains important structural patterns from the reference. CX Loss is particularly useful in reflection removal, style transfer, and image synthesis, as it helps maintain perceptual consistency while allowing flexibility in pixel arrangements.

In addition to traditional loss functions, some evaluation metrics are also directly used as loss terms. For instance, Zheng et al.~\cite{zheng2021single,wan2019corrn} directly incorporates PSNR, SSIM, and SI. PSNR ensures high-fidelity reconstruction, SSIM preserves structural similarity, and SI enhances overall structural consistency. By combining these metrics, the model improves reflection removal performance.

\section{Datasets \& Evaluation Metrics}
\label{sec:datasets}

\subsection{Data Acquisition}

The datasets for SIRR are a critical aspect of developing effective deep learning models. These datasets vary in size, image source, and the type of annotations provided. In general, they can be categorized into two main types: synthetic datasets and real-world datasets.

\subsubsection{Synthetic Datasets}
Synthetic datasets are created by simulating the reflection phenomenon using computer graphics techniques. This allows for precise control over various factors such as the intensity and blurriness of the reflection, as well as the presence of ghosting effects. Common methods for creating synthetic datasets include:

\textbf{Image Mixing:} Combining two images with different coefficients to represent the background and reflection layers.

\textbf{Reflection Blur:} Applying Gaussian blur to the reflection layer to mimic the out-of-focus effect.

\textbf{Brightness Adjustment:} Adaptively adjusting brightness and contrast to create realistic reflections.

\textbf{Physics-based Rendering:} Using physics-based methods to render reflections.

\subsubsection{Real-world Datasets}

Real-world datasets are captured using cameras in real-world environments. This provides more realistic and diverse data, but it also makes it more challenging to obtain accurate ground truth images. Common methods for creating real-world datasets include:

\textbf{Manual Glass Removal:} Capturing images with and without the glass to obtain ground truth.

\textbf{Raw Data Subtraction:} Subtracting the reflection from the mixed image in the raw data space.

\textbf{Flash/no-flash Pairs:} Capturing images with and without flash to exploit flash-only cues.

\textbf{Polarization:} Using polarization cameras to capture images with different polarization angles.

\textbf{Controlled Environments:} Capturing images in controlled environments with varying lighting, glass thickness, and camera settings.

\subsection{Current Public Datasets}

Several public datasets have been created to facilitate the development and evaluation of deep learning models for SIRR. These datasets vary in size, image source, and the type of annotations provided. Table~\ref{tab:dataset} summarizes the most important datasets for SIRR, including their usage (training, testing, or both), the number of image pairs, average resolution, and whether they collect from real or synthetic images.

\begin{table*}[ht]
\centering
\caption{Comparison of existing important reflection removal datasets. Syn: synthetic data, R: real data }
\resizebox{0.6\textwidth}{!}{
\begin{tabular}{clccccl}
\toprule
Dataset &  Year & Usage      & \begin{tabular}[c]{@{}c@{}}Pair \\ Number\end{tabular} & \begin{tabular}[c]{@{}c@{}}Average \\ Resolution\end{tabular}  &Real/Syn\\ \hline
CEIL~\cite{fan2017generic} &2017 & Train/Test& 7643/850& 224 x 224&Syn\\ 
 RID~\cite{wan2019corrn}  &2017& Train& 3250& 224 x 320&Syn\\ 
SIR$^2$~\cite{wan2017benchmarking}     &2017 & Test       & 454         & 540 x 400           &Real\\
Real~\cite{zhang2018single}          &2018 & Train/Test & 89/20       & 1152 x 930          &Real\\
Nature~\cite{li2020single}        &2020 & Train/Test & 200/20      & 598 x 398           &Real\\
CDR~\cite{lei2022categorized}        &2021 & Train & 1063      & 1542 x 1638           &Real\\
 SIR$^{2+}$~\cite{SIR2pami} & 2022& Train& 1700& 540 × 400&Real\\
 CID~\cite{wang2022background} & 2023& Train/Test& 4800/1200& 224 × 288&Syn\\
RRW~\cite{zhu2024revisiting}        &2024 & Train & 14952      & 2580 × 1460                   &Real\\ \bottomrule
\end{tabular}
}
\label{tab:dataset}
\end{table*}



Some of these datasets, such as Nature and Real, are relatively small and contain only real-world images with ground truth transmission layers. Others, such as SIR$^2$~\cite{wan2017benchmarking} and CEIL~\cite{fan2017generic}, are larger and include both synthetic and real-world images.  The SIR$^2$~\cite{wan2017benchmarking} dataset is particularly notable for its diversity, as it includes images with varying blur levels and glass thicknesses. The CEIL dataset, on the other hand, is designed to be more challenging, as it includes images with strong reflections. 

More recent datasets, such as CDR~\cite{lei2022categorized}, has been created to address the limitations of earlier datasets.  The CDR dataset is categorized according to reflection types and contains images with perfect alignment between the mixed and transmission images includes misaligned raw flash/ambient images. 

The largest and most recent dataset is RRW~\cite{zhu2024revisiting}, which contains over 14,950 high-resolution real-world reflection pairs. This dataset is particularly valuable for training deep learning models, as it provides a large number of images.

The choice of dataset depends on the specific application and the desired properties of the reflection removal algorithm. In general, larger and more diverse datasets are preferred, as they enable the training of more robust and generalizable models.

\subsection{Evaluation Metrics}

In this survey, we provide a comprehensive summary of evaluation metrics commonly employed in deep learning-based reflection removal. These metrics can be broadly classified into two categories: quantitative metrics and qualitative metrics.

\subsubsection{Quantitative Metrics}

Quantitative metrics are used to objectively measure the difference between the predicted transmission layer and the ground-truth transmission layer. Commonly used quantitative metrics include:

\begin{itemize}
\item \textbf{PSNR} (Peak Signal-to-Noise Ratio): Measures the difference between the predicted transmission layer and the ground-truth transmission layer. 
\item \textbf{SSIM} (Structural Similarity): Evaluates the similarity between the predicted and ground-truth transmission layers from three aspects: luminance, contrast, and structure. 
\item \textbf{MSE} (Mean Squared Error): Calculates the average squared difference between the predicted and ground-truth transmission layers. 
\item \textbf{Local MSE} (LMSE): Evaluates the local structure similarity by calculating the similarity of each local patch. 
\item \textbf{Normalized Cross Correlation} (NCC): Measures the correlation between the predicted and ground-truth transmission layers after normalizing their overall intensity. 
\item \textbf{Structure Index} (SI): Evaluates the structural similarity between the predicted and ground-truth transmission layers based on their covariance and variance. 
\end{itemize}

\subsubsection{Qualitative Metrics}

Qualitative metrics, on the other hand, are used to subjectively evaluate the visual quality of the predicted transmission images. One common qualitative metric is the \textbf{perceptual user study} \cite{zhang2018single}, where human users compare the predicted transmission images with the ground-truth images and rate their quality.

The choice of evaluation metrics depends on the specific application and the desired properties of the reflection removal algorithm. In general, a combination of quantitative and qualitative metrics is used to provide a comprehensive evaluation of the algorithm's performance.

\section{Discussion}
\label{sec:discussion}
\subsection{Challenges in current SIRR research}
One of the biggest challenges in SIRR research is the lack of large, high-quality training datasets that represent a variety of reflection types across different surfaces and lighting conditions. Reflection removal relies on supervised learning, which requires a well-labeled dataset with clear ground-truth images for training. However, creating or collecting such datasets is both time- and labor-consuming. Moreover, the absence of suitable test sets with real-world reflection scenarios presents another challenge. These test sets should include not only high-quality images but also a wide range of reflective surfaces, lighting conditions, and material properties (e.g., building glass, car window, smooth metal surface, rough metal surface, etc.). Without such comprehensive datasets, model evaluation remains limited and often unreliable when deploying into the real world.

Furthermore, the lack of datasets in SIRR research has also made exploring complex network architectures less valuable. With limited data, simpler models like UNet is already achieving state-of-the-art results, making the development of more complex models unnecessary and slowing progress in the evolution of SIRR network designs. These intricate architectures are prone to overfitting on small datasets, preventing them from reaching their full potential. Consequently, academic research in this field is stagnating, resulting fewer innovations in network design for SIRR.

In addition, the inherent complexity of the SIRR task itself compounds these challenges. Reflections vary in type, intensity, and interaction with the scene—some completely obscure the background, turning the task into a form of image inpainting. Such a wide variety makes it difficult to clearly define what a SIRR task should involve: is it purely about removing reflections, or does it also need to reconstruct missing background details? The lack of a comprehensive task definition hinders the development of consistent methodologies, reliable evaluation metrics, and meaningful comparisons. To make real progress, the academic community must come up with a clearer and more inclusive task definition that better captures the complexity of reflection removal and develop specific guidelines for handling different reflection scenarios.

\subsection{Future Directions}
One of the most pressing future directions in SIRR research is the creation of large-scale, high-quality datasets that cover a wide variety of reflection types, surfaces, and lighting conditions. Efforts should also focus on curating test sets that contain not only clean ground-truth images without reflections, but also cover real-world reflective materials (e.g., different types of glass and metals) and various environmental factors (e.g., lighting variations and reflections in challenging environments). Such test sets will enable more accurate training and better evaluation metrics, ultimately improving the generalization capabilities of SIRR models. To address the data scarcity issues, we appeal for collaborations between research institutions, industry, and the development of more powerful synthetic data generation methods.

On the other hand, future SIRR development could greatly benefit from the integration of advanced AIGC models. By utilizing large vision foundation models, researchers can enhance scene understanding and improve semantic reasoning, while large language models can provide deeper, descriptive insights into scene content and the relationships between reflections and transmissions.

Additionally, combining multimodal information (e.g., text descriptions, depth information, and semantic segmentation) will strengthen the reflection-transmission separation by leveraging complementary insights from multimodal resources. We believe that this fusion could result in more context-aware and precise reflection removal systems.

As mentioned earlier, clarifying the definition of SIRR task is another critical direction. The field needs to precisely define whether the goal is limited to reflection removal or extends to background reconstruction in cases of severe reflections. Establishing these clear boundaries will enable standardized evaluations and fair comparisons, ultimately leading to the development of more effective solutions.

\subsection{Limitations}
Our work has several limitations. First, due to the keywords and databases chosen for the search query, some related research may have been missed. In addition, research that is not published in English, exceeds the prescribed time frame, or does not provide enough technical information, were not included. Nevertheless, our paper provides a comprehensive analysis of the existing literature, based on a sample of 28 papers sourced from key venues. The aim of our review is to provide readers with a clear and rapid understanding of the key developments in SIRR field, including its current state, challenges, and future directions. Second, we did not include benchmark results in this paper, as different methods and datasets often have their own unique training strategies and evaluation process. To address this, we will develop a unified evaluation framework in the future, which will provide a fair platform to compare all publicly available datasets.

\section{Conclusion}
\label{sec:conclusion}
This article presents an overview of SIRR utilizing deep learning methods. We first describe the methodology used for our bibliographic search. Then, we provide a comprehensive analysis of research categorized by single-stage, two-stage, and multi-stage approaches. Next, we introduce the commonly used publicly available datasets and evaluation metrics. Finally, we discuss open challenges and future directions for SIRR research. We argue that with clearer definitions and larger-scale datasets, current deep learning methods can better realize their data-driven potential, leading to rapid advancements in the SIRR field in the near future.

\vfill\pagebreak



\let\oldthebibliography=\thebibliography
\let\endoldthebibliography=\endthebibliography
\renewenvironment{thebibliography}[1]{%
   \begin{oldthebibliography}{#1}%
     \setlength{\itemsep}{-.3ex}%
}%
{%
   \end{oldthebibliography}%
}

{
\footnotesize
\bibliographystyle{IEEEbib}
\bibliography{refs}
}
\end{document}